\documentclass[letterpaper, 10 pt, conference]{ieeeconf}

\IEEEoverridecommandlockouts
\usepackage{graphics} 
\usepackage{epsfig} 
\usepackage{times} 
\usepackage{amsmath} 
\usepackage{amssymb}  
\usepackage[table]{xcolor}
\usepackage[ruled,linesnumbered, noend]{algorithm2e}

\usepackage{soul}
\usepackage{cite}
\usepackage{comment}
\usepackage{multirow}
\usepackage{graphicx}
\usepackage{wrapfig}
\usepackage{soul, color}
\usepackage{wrapfig}
\usepackage{mathtools}
\usepackage{graphicx}
\usepackage{booktabs}
\usepackage{adjustbox}

\usepackage{caption}
\usepackage{subfigure}

\definecolor{LightCyan}{rgb}{0.88,1,1}
\definecolor{Gray}{gray}{0.9}

\newcommand{\ve}[1]{\mathbf{#1}} 

\title{Kinematic Optimization of a Robotic Arm for Automation Tasks with Human Demonstration}

\author{Inbar Meir, Avital Bechar and Avishai Sintov
\thanks{I. Meir and A. Sintov are with the School of Mechanical Engineering, Tel-Aviv University, Israel. e-mail: {\tt\small \{inbarb3, sintov1\}@tauex.tau.ac.il}.}
\thanks{A. Bechar is with the The Institute of Agricultural Engineering, Volcani Institute, Israel.}
\thanks{This research was partly supported by the Ministry of Science and Technology of Israel.} 
}

\begin{document}

\setlength{\belowdisplayskip}{2pt}
\setlength{\belowdisplayshortskip}{3pt}
\setlength{\abovedisplayskip}{2pt} 
\setlength{\abovedisplayshortskip}{3pt}
\setlength{\parskip}{0pt}


\maketitle
\thispagestyle{empty}
\pagestyle{empty}


\begin{abstract}
Robotic arms are highly common in various automation processes such as manufacturing lines. However, these highly capable robots are usually degraded to simple repetitive tasks such as pick-and-place. On the other hand, designing an optimal robot for one specific task consumes large resources of engineering time and costs. In this paper, we propose a novel concept for optimizing the fitness of a robotic arm to perform a specific task based on human demonstration. Fitness of a robot arm is a measure of its ability to follow recorded human arm and hand paths. The optimization is conducted using a modified variant of the Particle Swarm Optimization  for the robot design problem. In the proposed approach, we generate an optimal robot design along with the required path to complete the task. The approach could reduce the time-to-market of robotic arms and enable the standardization of modular robotic parts. Novice users could easily apply a minimal robot arm to various tasks. Two test cases of common manufacturing tasks are presented yielding optimal designs and reduced computational effort by up to 92\%. 
\end{abstract}

\section{Introduction}
\label{sec:introduction}

Robotic arms are the foundation of modern automation for manufacturing. They accommodate production lines and perform the majority of tasks such as assembly, machining, painting, welding and packaging \cite{Do2012, Beschi2019, Chutima2020}. While robotic arms are highly capable with multiple degrees-of-freedom (DOF), a robot is usually degraded to solely perform one simple repetitive task along the production line \cite{Cefalo2013}. These robotic arms are commonly standard off-the-shelve hardware which are redundant for the task at hand and, therefore, highly expensive having direct impact on the final product cost. Hence, many highly capable and expensive robots are purchased and integrated just to perform simple tasks. While simpler and cheaper robots can usually perform the same tasks, they are rarely available as an off-the-shelve product \cite{Bloch2018}. On the other hand, designing designated robots for each specific task requires complex and careful work by professional engineers. The process consumes a considerable amount of engineering time and may increase final product cost.
\begin{figure}
\centering
\includegraphics[width=0.8\linewidth]{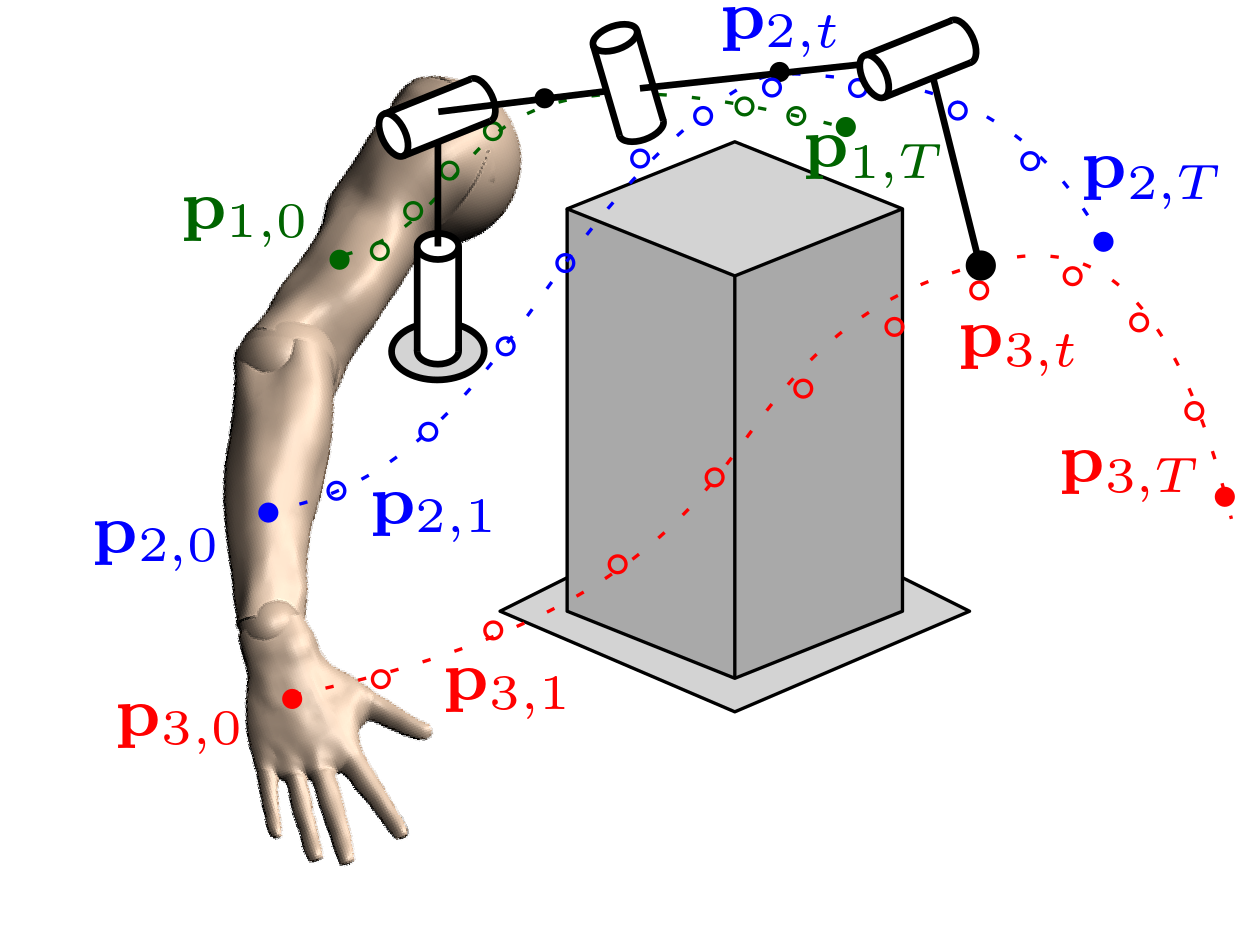}
\vspace{-0.6cm}
\caption{\small Illustration of a robot kinematics evaluated on a recorded task. In this example, the robot end-effector must follow a recorded path of the human hand (circular red markers) while its arm should follow the human arm markers (circular blue and green markers) to avoid obstacles. 
The dashed curves are the paths of the optimal robotic arm that best track the recorded human ones.}
\label{fig:track}
\vspace{-0.5cm}
\end{figure}

Some operational tasks are planned to comply with the natural motion of the human arm. Transitioning from manual work to robotic automation usually involves off-the-shelve robotic arms with kinematics that may not perform well in such environment. Hence, an optimally designed kinematic structure is expected to improve performance and reduce costs. Furthermore, while integrating the new robot, a path to be repeated over and over is usually pre-recorded by a technician while considering robot coordinates \cite{Perrollaz2012}. The recording is acquired through the teach pendant which provides non-intuitive and non-natural control abilities \cite{Schaal1996}. Consequently, the resulted path may be non-optimal to the task nor to the environment. Alternatively, explicit mapping of the environment is required in order to validate the ability of a candidate robot to execute a collision-free path \cite{Patel2015}.

In this work, we propose a novel concept of task-oriented robot design based on expert demonstration. We observe a human expert performing a task. The whole arm motion of the expert is recorded by tracking paths of markers from shoulder to hand (or grasped tool). Then, we formulate an optimization problem that searches for an optimal robotic arm that can accurately track the recorded task. In order to be able to avoid obstacles as the expert did, tracking includes the End-Effector (EE) of the robot as well as its entire kinematic chain as in Figure \ref{fig:track}. The optimization searches for the Denavit-Hartenberg (DH) parameters \cite{Patel2015} that define a robot kinematics. 

In this study, we propose the \textit{Robot Arm Particle Swarm Optimization} (RA-PSO) algorithm to efficiently solve the design problem. RA-PSO is a modified version of the known PSO method \cite{Kennedy1995} and is particularly aimed to optimize robotic arms based on recorded paths. RA-PSO has modified rules to fit the design space of robots and for efficient exploration of the valid search space. The search space in this problem is highly non-linear and non-continuous. 
We also observe the Computational Effort (CE) and compare it to the traditional PSO. Our approach distributes the search over different design variables to enable efficient exploration and low CE.

The method does not only provide an optimal robot design but also defines a collision free joint path to complete the task. Hence, no additional motion planning nor obstacle mapping are required. Indeed, an experienced practitioner can plan a more suitable path without considering demonstrations. However, a novice user may not have such capabilities and, therefore, the proposed approach reduces required engineering work and effort. A novice user can simply record an expert path and acquire both a robot design and a suitable motion. To reduce costs, this approach motivates the development of standard and modular robotic hardware, as proposed in \cite{Shi2006}, such that the novice user can rapidly assemble and operate custom arms for different tasks based on the optimization outputs. Consequently, the resultant arm could be low-cost, minimal and suitable for the environment and task with low engineering and computational efforts.




\section{Related Work}
\label{sec:related_work}

Many research studies have addressed the kinematic optimization of robotic arms while focusing mostly on increasing dexterity and workspace volume \cite{Vijaykumar1986,Stock2003}. Recent work in \cite{Zeiaee2019} used the Denavit-Hartenberg (DH) representation \cite{Murray1994} to define the design variables of an eight degrees-of-freedom (DOF) upper-limb rehabilitation exoskeleton. A constrained multi-objective optimization problem was formulated utilizing the Genetic Algorithm (GA) to search for the design variables that provide minimum size and maximum dexterity. The kinematic design optimization of an anthropomorphic robot using GA was proposed to maximize the multi-fingered precision grasping capability of the robot hand \cite{You2019}. While these addressed enhancing capabilities for general tasks, others have addressed a task-specific optimization problem \cite{Lum2004,Kuntz2018}. A kinematic optimization framework was also presented to find the DH parameters of a robot such that it can reach any task point in a defined task subset \cite{Patel2015}. 

Kinematic optimization solutions, such as the ones previously mentioned, use different performance measures to evaluate the kinematics of a robotic arm for general tasks \cite{Rastegar1990}. A common measure is the Manipulability ellipsoid 
that approximates the distance to singular configurations \cite{Vahrenkamp2012}. Similarly, the Global Isotropy Index (GII) measures the ratio between the minimum and maximum singular values of the Jacobian \cite{Stocco1998}. 
These, however, do not consider tracking accuracy for task-specific paths.

Since cost functions that reflect the kinematics of the robot arm are non-linear, gradient-based techniques will most likely fail to find global solutions and, therefore, are not common \cite{Zeiaee2019}. On the other hand, nature-inspired meta-heuristic approaches (e.g., PSO, Artificial Bee Colony, Ant Colony Optimization) are more capable in searching for the global minimum in complex problems \cite{GiladiSintov2020}. For instance, Simulated Annealing (SA) was used in the design of a surgical robot to optimize anatomical visibility \cite{Kuntz2018}. GA was used to optimize the design of a manipulator with the GII measure \cite{Khatami2002}. Similarly, PSO was used to identify the optimal cable routing for a cable-driven manipulator \cite{Bryson2016}. We observe the performance of these methods for robot design.

Human demonstrations have been used in robotics for many applications such as augmenting reinforcement learning \cite{Zhu2018}, motion re-targeting \cite{Ayusawa2017}, imitation learning \cite{Laura2020}, motion planning \cite{Garcia2019} and computational design \cite{Ceylan2013, Coros2013}. In addition, task-centric optimization was introduced to select pose configurations of an existing robot for efficient assistive tasks based on sampled human poses \cite{Kapusta2016}. However and to the best of the author's knowledge, no attempt has been made to optimize the kinematics of a robotic arm for a whole arm tracking of a human demonstrated task. A closely related work proposed the synthesis of an arm by fitting EE poses on a set of points \cite{Gracia2006}. Similarly, kinematic synthesis was proposed such that the EE of an under-actuated robotic arm would track some path \cite{Shirafuji2019}. However, the derivative of the path must be available at any point for generalized differential Inverse Kinematics (IK). Such derivative cannot be acquired qualitatively from a recorded demonstration. Furthermore, optimization has been attempted to solely decide joint displacements and, did not observe human demonstrations. In addition, having an EE track a path does not ensure that it will be able to prevent collisions with the arm links. Hence, we consider the path of the entire arm in order to avoid obstacles in the environment and provide a complete solution.

\section{Methodology}

\subsection{Optimization objective} 

A task is given in which an $n$ degrees-of-freedom (DOF) robotic arm must move in an industrial environment with obstacles. The task could be moving an object from one pose to another (e.g., packaging) or tracking a path (e.g., welding). The assumption is that the environment is static. The human path is recorded once and, therefore, dynamic obstacles cannot be considered. 
We aim to find an optimal robotic arm that can perform the task. Without loss of generality, we consider only rotary joints. An optimal robot is the one that 
provides best accuracy during task execution. We consider position errors for the entire arm from base to end-effector so that arm distance from obstacles can be maximized. In this work, we address the optimization problem of the kinematic properties while non-optimal dynamic parameters are commonly compensated using slow motions and control techniques. When the system and task environment are much larger than human capabilities, a scaled mock-up can be used with the proposed approach to compute an optimal robot and later be scaled-up. 



\subsection{Approach}

As discussed previously, we consider a task in a static environment. Hence and in light of the above objective, we seek for a robotic arm with minimal DOF that can accurately track a single path. The single path, in this work, is a path recorded by a human demonstrator. We assume that a path recorded by a human expert is the shortest while maximizing the distance from obstacles. The robot is modeled with thin segments and, therefore, we assume that the recorded paths are far enough from obstacles. Hence, the better a robot is able to accurately track the path, the less it is sensitive to positioning errors and risk of collisions.

Given a task, the motion of the arm of a human expert is recorded. Let $m$ be the number of tracked markers on the arm, a recorded task is given by $m$ paths
\begin{equation}\label{eq:robotPath}
 \mathcal{P} = \{P_1,P_2,\ldots,P_m\}
\end{equation}
where $P_i=\{\ve{p}_{i,0},\ldots,\ve{p}_{i,T}\}$ is a path of the $i^{th}$ marker relative to the shoulder. $\ve{p}_{i,t}\in\mathbb{R}^3$ is the spatial coordinates of the $i^{th}$ marker at time $t$ (Figure \ref{fig:track}). We note that $P_m$ is the path of the human hand, which implies about the desired path of the end-effector. We also record a set $\mathcal{V}=\{\ve{u}_{0},\ldots,\ve{u}_{T}\}$ where elements in $\ve{u}_{t}\in\mathbb{R}^3$ are the Euler angles of the hand or tool at time $t$. In a case where the recorded paths are not smooth due to hand shaking or vibration, some filtering can be applied. The recorded task can now be used to evaluate the ability of some robot arm to accurately track $\mathcal{P}$. 


\begin{figure}
\centering
\begin{tabular}{cc}
     \includegraphics[width=0.49\linewidth]{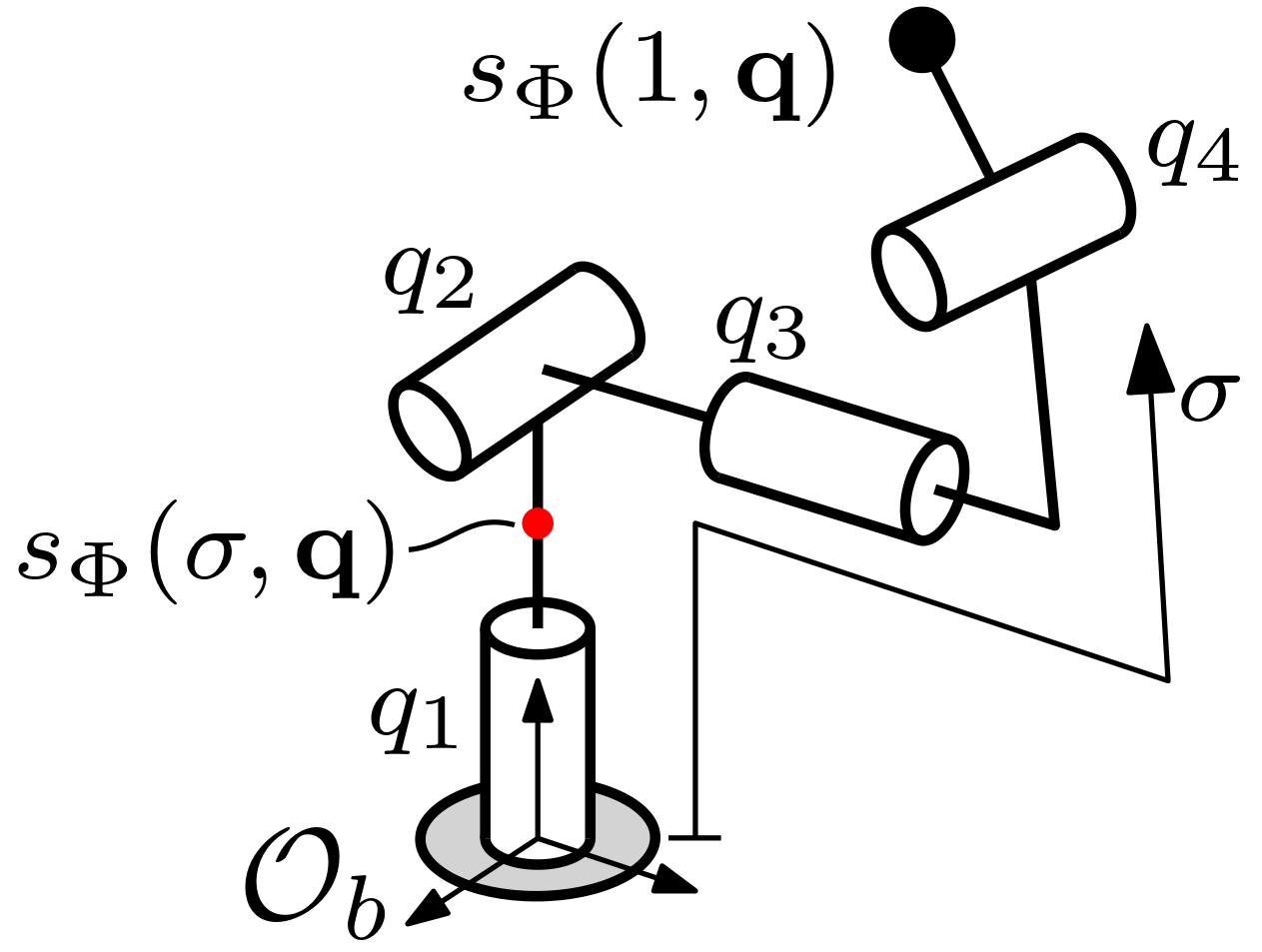} &
    \includegraphics[width=0.45\linewidth]{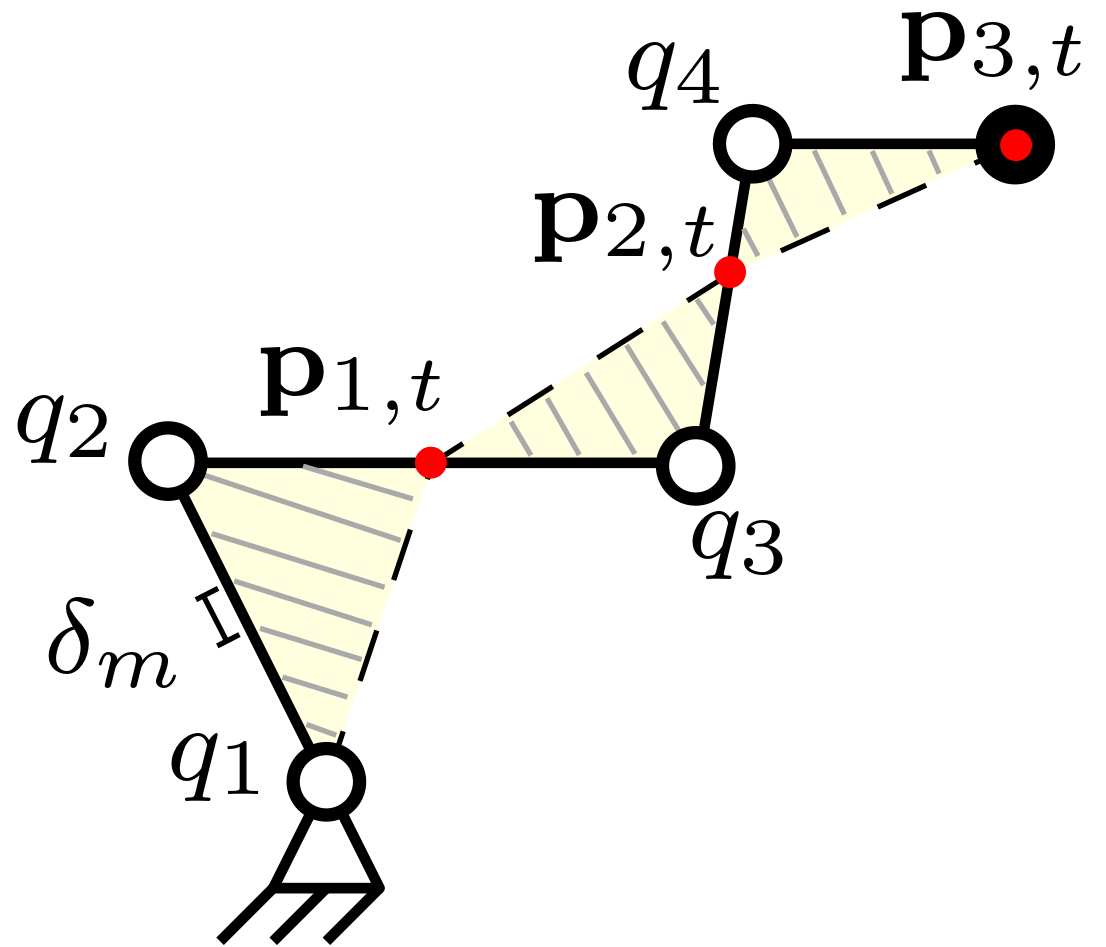} \\
    (a) & (b)\\
\end{tabular}
\caption{\small (a) Illustration of a robot parameterized by $\Phi$ where any point along it can be represented by $\ve{s}_\Phi(\sigma,\ve{q})$. Points $\sigma=0$ and $\sigma=1$ are the base and end-effector points, respectively. (b) Area to minimize (yellow) for a better fit between the robot arm (solid lines) and lines (dashed) formed by the recorded markers (red).}
\label{fig:RobotVec}
\vspace{-0.4cm}
\end{figure}

\subsection{Robot design variables} 
\label{sec:RobotConfigOpt}

To optimize the design of an $n$-DOF robotic arm, we first define the variables that describe its kinematics. The kinematics of a robot can be defined by a set of coordinates described by the Denavit-Hartenberg (DH) method \cite{Murray1994}. In DH, the forward kinematics of an arm is represented by the product of a set of canonical homogeneous transformation matrices $A_i\in SE(3)$ where $i\in\{1,\ldots,n\}$. Transformation matrix $A_i(a_i,\alpha_i,\theta_i,d_i)$ defines the position and orientation of the coordinate frame of joint $i$ relative to frame $i-1$. The four parameters $a_i$, $\alpha_i$, $\theta_i$ and $d_i$ are standard DH geometric quantities. Matrix $A_i$ incorporates only one DOF and, thus, depends only on one variable $q_i$. This variable would be the joint angle $q_i=\theta_i$  for a revolute one. The remaining three are constants and part of the design of the robot. Let $\phi_i\in\mathbb{R}^3$ be a parameterization vector that encodes the parameters of transformation from frame $i$ to frame $i-1$. Hence, $\phi_i$ is given by 
\begin{equation}
    \label{eq: jointV}
    \phi_i =(\alpha_i, a_i, d_i).
\end{equation}
We define $\ve{q}=(q_1,\ldots,q_n)\in\mathcal{C}$ as the vector of joint values of the arm where $\mathcal{C}\subset\mathbb{R}^n$ is the configuration space of the robot. Vector $\Phi\in\Omega$, where $\Omega\subseteq\mathbb{R}^{3n}$, is the concatenation of all $\phi_i$ given by $\Phi = (\phi_1,..,\phi_{n})$ and is the encoding of the constant parameters.
Consequently, the product
\begin{equation}
    \label{eq:Phi}
     A_{ee}(\ve{q},\Phi)=\prod_{i=1}^n A_i(q_i,\phi_i)   
\end{equation}
is the robots forward kinematics that express the position and orientation of its EE with regards to the base frame $\mathcal{O}_b$. Vector $\Phi$ in \eqref{eq:Phi} fully-defines the design of a robot and, therefore, $\Omega$ is the search space for the optimization problem.

\subsection{Allocate robot tracking point}

Given a robot defined by $\Phi$, let $\ve{s}_\Phi:[0,1]\times\mathcal{C}\to\mathbb{R}^3$ be a map such that $\ve{s}_\Phi(0,\ve{q})=\ve{0}$ is the position of the robot base and $\ve{s}_\Phi(1,\ve{q})$ is the position of its EE. 
Map $\ve{s}_\Phi$ is computed using the forward-kinematics described in Section \ref{sec:RobotConfigOpt}. Hence, $\ve{s}_\Phi(\sigma,\ve{q})$ with $\sigma\in[0,1]$ is a parametric function providing the position of any given point $\sigma$ along a one-dimensional representation of the arm at configuration $\ve{q}$ as seen in Figure \ref{fig:RobotVec}a. Function $\ve{s}_\Phi$ is non-smooth due to transitions between links at the joints. We also define function $\ve{b}_\Phi(\ve{q}):\mathcal{C}\to\mathbb{R}^3$ which outputs the Euler angles of the EE for robot $\Phi$ at configuration $\ve{q}$.

\subsection{Robot temporal fitness}

A recorded task \eqref{eq:robotPath} is composed of a sequence of time frames. For each individual frame, we seek to evaluate the fitness of a given robot to the corresponding recorded points. A \textit{temporal fitness} at time $t$ is defined to be the weighted root-mean-square error (wRMSE) between points $\ve{p}_{1,t},\ldots,\ve{p}_{m,t}$ and the robot. Therefore, the wRMSE at time $t$ for robot configuration $\ve{q}$ is given by
\begin{align}
\label{ep:Objective function-3}
   g_\Phi(t,\ve{q}) =\frac{1}{m}  \sqrt{ \sum_{i=1}^m w_i\|\ve{s}_\Phi(\sigma_i,\ve{q})-\ve{p}_{i,t}\|^2 }+\\+w_0 \|\ve{b}_{\Phi}(\ve{q})-\ve{u}_{t}\|\nonumber
\end{align}
where parameter $\sigma_i$ associates $\ve{p}_{i,t}$ to the closest point on the robot. Scalars 
$\{w_0,\ldots,w_m\}\in\mathbb{R}^+$ with $\sum_0^m w_i=1$ are user-defined weight values that prioritize the position importance of the robot segments relative to the recorded path. In a non-cluttered environment, for instance, the position of the end-effector may have higher significance compared to other links. On the other hand, when working in cluttered environments, the body link positions have higher importance.




The temporal robot fitness $G_t$ at time $t$ for a robot represented by $\Phi$ is given by
\begin{equation}\label{eq:temporal robot fitness}
\begin{aligned}
G_{\Phi,t} = \min_{\ve{\sigma}, \ve{q}_t} \quad & g_\Phi(t,\ve{q}_t) \\
\textrm{s.t.} \quad & \ve{q}_t \in \mathcal{C}_f\\
                    & \|\ve{q}_t-\ve{q}_{t-1}\|\leq \epsilon \\
  & \sigma_j < \sigma_k,~~ \forall j<k    \\
\end{aligned}
\end{equation}
where $\ve{\sigma}=(\sigma_1,\ldots,\sigma_m)^T$, $\ve{q}_t\in\mathcal{C}$ is the joint configuration at time $t$ and $\mathcal{C}_f\subset\mathcal{C}$ is the feasible configuration subspace to be discussed below. The second constraint enforces joint path continuity by requiring that two consecutive configurations have a distance smaller than $\epsilon>0$. Hence, the initial guess for optimization problem \eqref{eq:temporal robot fitness} would be $\ve{q}_{t-1}$. Optimization problem \eqref{eq:temporal robot fitness} minimizes the wRMSE of the euclidean distances for all recorded markers from the robot for a single time frame. The solution of problem \eqref{eq:temporal robot fitness} is, in fact, the solution of a multi-point Inverse Kinematics. Next, we use this formulation to calculate the fitness along an entire path.

\subsection{Robot path fitness}
Given path set $\mathcal{P}$, the \textit{path fitness} $f(\Phi)$ of a query robot design $\Phi$ is the mean of the temporal fitnesses given by \eqref{eq:temporal robot fitness} for all time frames $t\in[0,T]$. Formally, a path fitness for robot $\Phi$ is defined as
\begin{equation}
    \label{eq:f_Phi}
    f(\Phi) = \frac{1}{T}\sum_{t=0}^T G_{\Phi,t}.
\end{equation}
Since we require continuous joint motion and we fix the initial pose of the end-effector based on the task, \eqref{eq:f_Phi} is computed sequentially from $t=0$.

Minimizing $f(\Phi)$ reduces the distance between the robot and recorded markers. However, such minimization is not sufficient since the output may yield a long arm that over-fits the markers as seen on Figure \ref{fig:RobotVec}b. That is, the path fitness can be zero while the arm is very long. Hence, we also minimize the area $E(\Phi)$ (yellow region in the figure) between robot links and represented markers. $E(\Phi)$ is computed as follows. Recall that, at each time frame, the arm is separated into $m$ parts by $\sigma$ from the solution of \eqref{eq:temporal robot fitness}. We divide each part into segments of length $\delta_m$ along the robot axis. For each segment, we calculate the distance from the line formed by the two closest markers. $E(\Phi)$ is the mean distance along the robot length and at all time frames. Note that minimizing $E(\Phi)$ indirectly also minimizes robot length. 


Finally, the optimal robotic arm $\Phi^*$ that can track path $\mathcal{P}$ most accurately is the one with the minimal \textit{robot fitness} value and is the solution of
\begin{equation}\label{eq:robot path fitness }
\begin{array}{lrrclcl}
\Phi^* = &   \displaystyle arg\min_{\Phi} & \multicolumn{3}{l}{\lambda_f f(\Phi) + \lambda_E E(\Phi)} \\
&\textrm{s.t.} & \Phi \in \Omega_f\\
\end{array}
\end{equation}
where $\lambda_f>0$ and $\lambda_E>0$ are weighting values. 
Hence, we wish to minimize the tracking accuracy, link lengths and spatial link movements. Minimization of link lengths will result in a smaller and cheaper arm, along with lower joint loads, i.e., not requiring high-torque actuators. It is important to note that the higher the dimensionality of $\Phi$, i.e., more DOF, it is more likely for the robot to better track the path. However, we seek for a design comprising of minimal DOF. Subset $\Omega_f\subset\Omega$ is the allowed robot design search space defined in the next section.

\subsection{Boundaries and Constraints}\label{se:con}
We now define the joint and design search spaces, $\mathcal{C}_f$ and $\Omega_f$, respectively. The allowed joint search space $\mathcal{C}_f$ in \eqref{eq:temporal robot fitness} ensures proper physical movement of a query robot configuration and is defined by the joint limits 
\begin{equation}
  \mathcal{C}_f=\{ \ve{q}\in\mathcal{C}|~\ve{q}_{min}<\ve{q}<\ve{q}_{max}\}
\end{equation}
where $\ve{q}_{min}$ and $\ve{q}_{max}$ are the minimum and maximum actuators angle limits, respectively. For any query robot $\Phi$, the world coordinate frame is positioned at the base. The EE coordinate frame is fixed to the last robot link. The static values for the DH parameters in \eqref{eq: jointV} are limited according to
\begin{equation}
\label{eq:alpha_limit}
    \alpha_i\in[\alpha_{min},\alpha_{max}],~a_i\in[0,a_{max}]\text{ and } d_i \in [0,d_{max}]
\end{equation}
where $\alpha_{min},~\alpha_{max},~a_{max}$ and $d_{max}$ are pre-defined user variables. Furthermore, we constrain any two consecutive joints to be non-concentric in order to avoid redundancy. Such constraint assists in tightening the search sub-space. 
A robot solution is considered invalid and omitted if its total length $L$ does not satisfy
\begin{equation}\label{eq: length}
    L_{min}<L<L_{max}
\end{equation}
where $L_{min}$ and $L_{max}$ are user-defined values. This constraint also aims to filter-out lengthy designs early on. Moreover, if the tested robot configuration cannot reach EE position at $t=0$ with accuracy of 1~mm, the rest of the path is not calculated. As stated above, the valid search space for $\ve{q}_{t+1}$ depends on the previous solution $\ve{q}_{t}$. However, the valid search space for  $\ve{q}_0$ is defined by the joint angular limits. Hence, the robot can be initialized from different joint values and the rest of the path will be defined accordingly. Nevertheless, preliminary tests show high repetition for $\ve{q}_{0}$.

Let $\Omega_l\subset\Omega$ be the set of design configurations that satisfy \eqref{eq:alpha_limit}-\eqref{eq: length} and let $\mathcal{Q}_\Phi$ be the workspace of robot $\Phi$ such that $\ve{s}_\Phi(\sigma,\ve{q})\in\mathcal{Q}_\Phi$ for any $\ve{q}\in\mathcal{C}_f$ and $\sigma\in[0,1]$. Consequently, a valid robot configuration must lie within sub-set $\Omega_f\subset\Omega$ defined as 
\begin{equation}
    \label{eq:worckspace}
    \Omega_f=\{ \Phi\in\Omega | ~\Phi\in\Omega_l,~\mathcal{P}\subset\mathcal{Q}_\Phi \}.
\end{equation}
Subset $\Omega_f$ contains all valid robot configurations. 

\begin{figure*}[h]
    \centering
    \begin{tabular}{cc}
        \includegraphics[width=0.45\textwidth]{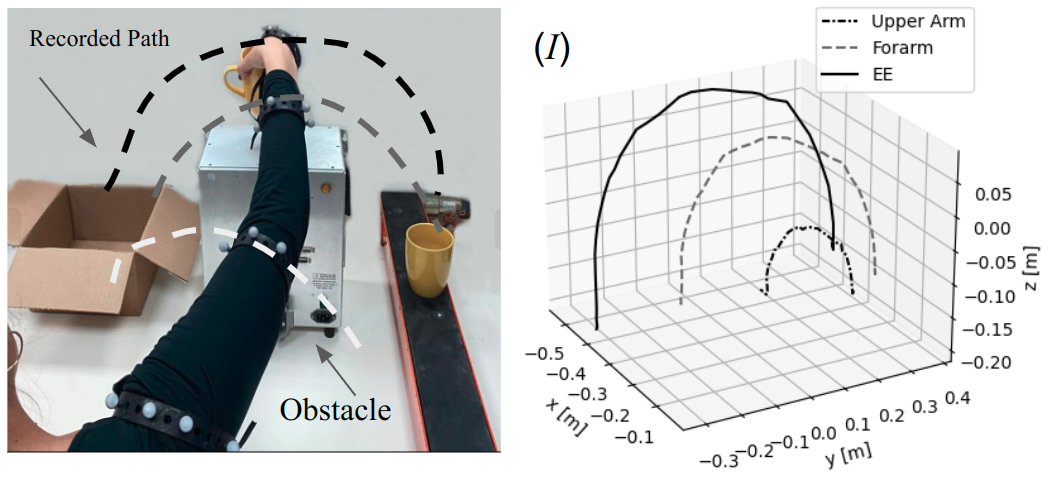} & \includegraphics[width=0.45\textwidth]{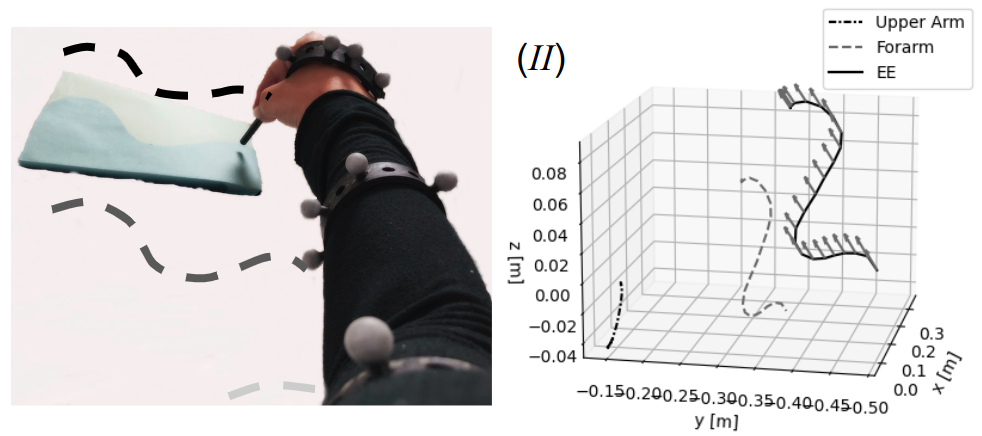} \\
    \end{tabular}
    \vspace{-0.3cm}
    \caption{\small Recording of a human arm path in an industrial (Scenario \textit{I}) packaging and (Scenario \textit{II}) welding. A robotic arm is to be optimized to accurately track the recorded path. The end-effector of the robot must follow the path of the hand markers while its arm should follow the human arm markers to avoid obstacles. For Scenario \textit{II}, the EE must also track the orientation of the human hand.}
    \label{fig:scenes}
    \vspace{-0.4cm}
\end{figure*}

\subsection{Optimization algorithm}
\label{sec:rapso_opt}
The definition of the robot fitness and constraints can now be used to search for the optimal design $\Phi^*$ of a robot. As discussed in Section \ref{sec:introduction}, kinematic optimization problems are commonly highly non-linear and cannot be efficiently solved using gradient-based techniques. Hence, we employ a meta-heuristic search approach.

Our proposed algorithm is based on the PSO, a meta-heuristic optimization algorithm. In PSO, we randomly sample $N$ particles $\mathcal{K}=\{\Phi_1,\ldots,\Phi_N\}$ from a uniform distribution in $\Omega_l$. Each particle $\Phi_j$ is evaluated yielding robot fitness $r_j$ and the corresponding path $Q_j=\{\ve{q}_0,\ldots,\ve{q}_T\}$ by solving the cost function in \eqref{eq:robot path fitness }. We store and update a personal best position vector $bp_j$ for particle $j$ based on robot fitness. The global best position for all the particles in $\mathcal{K}$ is stored in $\Phi^*$ with fitness $r^*$. Each particle is then updated according to $\Phi_{j}\leftarrow\Phi_{j}+\ve{v}_{j}$ where $\ve{v}_j$ is updated by
\begin{equation}\label{eq:PSO_update}
     \ve{v}_{j}\leftarrow w~\ve{v}_{j} + \beta_1 c_1(bp_j -\Phi_{j})+\beta_2 c_2(\Phi^{*} - \Phi_{j})),
\end{equation}
constants $w$, $c_1$ and $c_2$ are tunable parameters, and $\beta_1$ and $\beta_2$ are random numbers in $[0,1]$. The particles are iteratively updated for $M$ iterations.
Preliminary results have shown that the standard update rules of PSO face difficulties converging to a near optimal solution due to the large hyper-volume and non-linear behavior of $\Omega$. 

In order to improve the search, we propose the RA-PSO algorithm, a designated variant of PSO, designed to address the properties of search space $\Phi$. RA-PSO maximizes exploitation of particles within $\Omega_f$ while avoiding missing-out improved solutions nearby. In order to improve local search, we include two modifications in the update rules of PSO. First, we distinguish between particles currently in $\Omega_f$ and ones that are not. 
Particles not in $\Omega_f$ are updated according to the standard PSO which cause aggressive change in the candidate robot kinematics to explore new regions. On the other hand, we update a particle $\Phi_{j}$ that is within $\Omega_f$ in a significantly lower rate such that its new position is nearby for local refinement and effective exploitation. The velocity $\ve{v}_{j}$ of an agent in $\Omega_f$ is updated according to
\begin{equation}
    \ve{v}_{j} \gets \ve{v}_{j} + \texttt{random}(c_{min},c_{max}) \cdot \phi_{range}
\end{equation}
where $c_{min},c_{max}$ are random number bounds and $\phi_{range}$ is the range of the corresponding DH feature in $\Omega_f$.

The second modification improves local search in $\Omega_f$. Since $\Omega$ is large,  high-dimensional and non-linear, we explore different dimensions of $\Omega$ by their physical representation. That is, we separately update DH angular (i.e., $\alpha_i$) and length (i.e., $a_i$ and $d_i$) variables. In such way, the algorithm is able to better explore subspaces of $\Omega_f$. Let $D\in\mathbb{N}$ be the update frequency of the angular variables. Then, angular variables in $\Phi_j$ are updated every $D$ iterations. This method assists in testing various link lengths with the same angular variables and better improve the exploration of $\Omega_f$. 


\begin{table}[]
\centering
\caption{\small Optimization parameters}
\label{tb:opt_param}
\begin{adjustbox}{width=0.95\linewidth}
\begin{tabular}{|c|c|c|c|}\hline
\multicolumn{4}{|c|}{User-defined parameters}\\\hline
$l$ ~ 0.7~m  & $L_{min}$  ~ 0.6~m  & $L_{max}$  ~ 1.2~m & $\epsilon$ ~ $10^\circ$ \\
$a_{max}$ ~ $0.5$~m  & $d_{max}$ ~ $0.5$~m & $\alpha_{min}$~$-90^\circ$ & $\alpha_{max}$~$90^\circ$  \\
$q_{max}$ ~ $180^\circ$  & $q_{min}$ ~ $-180^\circ$ & $\delta_m$ ~ 10 mm &\\\hline
\multicolumn{4}{|c|}{Optimization hyper-parameters}\\\hline
$w$ ~ $0.8$ & $c_1$ ~ $0.4$ & $c_2$ ~ $0.6$ & $N ~ 400$ \\
$c_{max}$ ~ $0.5$ & $c_{min}$ ~ $-0.5$ & $M ~ 200$ &   \\\hline
\multicolumn{4}{|c|}{Scenario \textit{I} - Optimization weights}\\\hline
$w_0$ ~ 0 & $w_1$ ~ $1/6$  & $w_2$  ~ $1/3$ & $w_3$  ~ $1/2$\\
$\lambda_f$ ~ 15 &  $\lambda_E$ ~ 5 & &\\\hline
\multicolumn{4}{|c|}{Scenario \textit{II} - Optimization weights}\\\hline
$w_0$ ~ 0.2 & $w_1$ ~ $0$  & $w_2$ ~ $0.1$ & $w_3$  ~ $0.7$  \\
$\lambda_f$ ~ 15 & $\lambda_E$ ~ 5 & &\\\hline
\end{tabular}
\end{adjustbox}
\vspace{-0.4cm}
\end{table}


\section{Experiments}

\subsection{Setup and apparatus}

Using a motion capture system, we have recorded a human arm preforming desired tasks. During recording, four bands are places on the shoulder, upper arm, forearm and hand. Each band has several reflective markers considered together as a rigid body. Hence, the motion capture system provides real-time data stream of spatial positions of the bands. Data stream may also include orientation of the bands and the hand band in particular. The shoulder is considered as the base position. The desired robot motion $\mathcal{P}$ is, therefore, the paths of the $m=3$ distal bands relative to the shoulder band. All computations were implemented in Python on an Intel-Core Intel Xeon Gold 6130 Ubuntu machine with 12$\times$32GB of RAM. Optimization algorithms are based on the Opytimizer package \cite{rosa2019opytimizer}. 
\begin{table}[]
\centering
\caption{\small Optimization performance for solving robot temporal fitness \eqref{eq:temporal robot fitness} using various meta-heuristic algorithms}
\label{tb:score}
\begin{adjustbox}{width=\linewidth}
\begin{tabular}{c cccc c}\hline
      & \multicolumn{4}{c}{\textbf{Mean} $G_{\Phi,t}$ (mm)} & \textbf{Mean comp.} \\ 
     $n$ & 3             & 4                & 5             & 6               & \textbf{time (s)}  \\ \hline

GA    & 49.0 $\pm$25.1                              & 72.9$\pm$50.9             & 48.5$\pm$41.7                               & 36.1$\pm$17.6            & 10.1  \\ 
\rowcolor{Gray}

PSO   & 49.3 $\pm$25.0                              & 71.9 $\pm$52.1           & 48.7$\pm$42.9                               & 32.7 $\pm$17.9          & 5.7    \\ 

WOA   &  53.1$\pm$42.8                              & 74.7 $\pm$50.0           & 71.8$\pm$9.9                               & 40.9 $\pm$19.5           & 5.8    \\ 

ES    & 48.9 $\pm$25.1                               & 71.9$\pm$51.8             & 49.4$\pm$42.8                               & 35.9 $\pm$18.4           &  8.7 \\ 

ABC   & 48.7 $\pm$25.2                              & 69.5 $\pm$53.2            & 44.4$\pm$43.6                               & 29.5 $\pm$18.4           & 17.5 \\ 

SLSQP & 179.2$\pm$33.5                              &207.4 $\pm$34.3             & 203.9$\pm$40.7                               & 214.8$\pm$30.9            & 0.021   \\ 

SA    & 51.0 $\pm$24.6                             & 82.4$\pm$51.7             & 63.0$\pm$45.3                               & 54.9$\pm$19.6            & 11.5   \\ \hline
\end{tabular}
\end{adjustbox}
\vspace{-0.4cm}
\end{table}

\subsection{Test cases}

Two test scenarios, seen in Figure \ref{fig:scenes}, are considered. In packaging Scenario \textit{I}, an item is picked-up from a conveyor belt, manipulated around an obstacle and placed within a cardboard box; and a welding Scenario \textit{II} where a tool must trace a path across an object. The figures also show the recorded paths $\mathcal{P}$ of the three distal bands relative to the shoulder band. In Scenario \textit{II}, we constrain the robot EE to also trace the orientation of the human hand. In both scenarios, one repetition of the human hand performing the task was recorded. 
The user-defined parameters and optimization hyper-parameters used for both tests cases are shown in Table \ref{tb:opt_param}. The hyper-parameters were chosen based on preliminary analysis.

\subsection{Results \& Discussion}

A comparison analysis was conducted on seven optimization algorithms to solve robot temporal fitness $G_{\Phi,t}$ in \eqref{eq:temporal robot fitness}. The optimization algorithms include PSO, Artificial Bee Colony (ABC), Evolution Strategies (ES), Whale Optimization Algorithm (WOA), Genetic Algorithm (GA), Simulated Annealing (SA) and Sequential Least Squares Programming (SLSQP) \cite{Dokeroglu2019}. We note that RA-PSO is proposed for minimizing path fitness while these are benchmarked to solve only for the temporal fitness. The results of each algorithm are averaged over ten different robot configurations in three different time instants along $\mathcal{P}$. Results are summarized in Table \ref{tb:score}. Presented time results were computed on the same computer. First, SLSQP does not converge well and is, therefore, not suitable for this problem. All other algorithms converge relatively similar with small variations about the mean temporal fitness. In particular, PSO provides somewhat good convergence with the lowest computation time. Therefore, the temporal fitness is further solved using PSO.

Furthermore, we provide mean (over 30 repetitions) robot fitness results over all $n$ for solving problem \eqref{eq:robot path fitness } with four different algorithms: ABC, GA, PSO and SA. The results, presented in Table \ref{tb:score_diffrentAlg}, show that PSO provides best fitness and computation time. This provides motivation for pushing forward with PSO to improve convergence.

We next observe the impact of update frequency $D$ in RA-PSO on fitness and computational effort (CE) for both scenarios. 
While both PSO and RA-PSO share the same complexity, their CE is significantly affected by two factors. First, the more valid agents in $\Omega_f$ results in more cost evaluations. If the two algorithms reach the same fitness while one requires less agents, its effort is lower. Second, more iterations to convergence requires more computations. The CE is, therefore, $\bar{N}_v\cdot W$ where $\bar{N}_v$ is the mean number of valid agents and $W$ is the number of iterations to convergence \cite{Mousakazemi2020}. An algorithm with a lower CE is more computation efficient. Mean fitness and CE improvement results (over 30 repetitions) with regards to $D$ are shown in Table \ref{tb:score_RAPSO}. Negative results indicate an improvement compared to PSO. While fitness improvement marginally occurs in some cases, RA-PSO significantly improves CE over all $D$ for both scenarios. Results show the ability of RA-PSO to better explore $\Omega_f$ with less agents and iterations, while also maintaining low fitness. 

\begin{table}[]\caption{\small Optimization performance over all $n$ for solving robot fitness \eqref{eq:robot path fitness } using various meta-heuristic algorithms}
\label{tb:score_diffrentAlg}
\centering
\begin{adjustbox}{width=0.65\linewidth}
\begin{tabular}{l cccc}
\hline
\textbf{Alg.} & \textbf{ABC} & \textbf{GA} & \textbf{PSO} & \textbf{SA} \\\hline
Mean $\Phi^*$  & 0.64 & 0.70 & \cellcolor[HTML]{D9D9D9}0.51 & 0.71\\
Comp. time (h) & 3.59 & 4.00 & \cellcolor[HTML]{D9D9D9}0.91 & 1.28 \\
\hline
\end{tabular}
\end{adjustbox}
\end{table}
\begin{table}[]\centering
\caption{\small Optimization performance of RA-PSO for solving robot fitness \eqref{eq:robot path fitness } }
\label{tb:score_RAPSO}
\begin{adjustbox}{width=\linewidth}
\begin{tabular}{c c cccc cccc }
\hline
& & \multicolumn{4}{c}{$\Phi^*$ \textbf{improvement (\%)}} & \multicolumn{4}{c}{\textbf{CE improvement (\%)}} \\ 
& & \multicolumn{4}{c}{D} & \multicolumn{4}{c}{D} \\ \hline
n & Sc. & 2  & 3  & 4  & 5 & 2 & 3 & 4 & 5 \\ \hline
\cellcolor[HTML]{FFFFFF} & \cellcolor[HTML]{FFFFFF}\textit{I} & {\cellcolor[HTML]{D9D9D9}-10.5} & {\cellcolor[HTML]{D9D9D9}-15.6} & {\cellcolor[HTML]{D9D9D9}-5.9} & \cellcolor[HTML]{D9D9D9}-7.4  & {\cellcolor[HTML]{D9D9D9}-90.6} & {\cellcolor[HTML]{D9D9D9}-92.0} & {\cellcolor[HTML]{D9D9D9}-90.4} &\cellcolor[HTML]{D9D9D9} -88.9 \\ 
\multirow{-2}{*}{\cellcolor[HTML]{FFFFFF}3} & \cellcolor[HTML]{FFFFFF}\textit{II}   & {\cellcolor[HTML]{FFFFFF}3.0}   & {\cellcolor[HTML]{FFFFFF}9.8}   & {\cellcolor[HTML]{FFFFFF}0.8}  & \cellcolor[HTML]{FFFFFF}5.0  & {\cellcolor[HTML]{D9D9D9}-89.9} & {\cellcolor[HTML]{D9D9D9}-91.8} & {\cellcolor[HTML]{D9D9D9}-91.0} & \cellcolor[HTML]{D9D9D9}-88.9 \\ \hline
                                            & \textit{I}                         & {1.4}                           & {0.9}                           & {\cellcolor[HTML]{D9D9D9}-0.4}                         & \cellcolor[HTML]{D9D9D9}-7.0                         & {\cellcolor[HTML]{D9D9D9}-66.7} & {\cellcolor[HTML]{D9D9D9}-75.0} & {\cellcolor[HTML]{D9D9D9}-64.2} & \cellcolor[HTML]{D9D9D9}-66.7 \\
\multirow{-2}{*}{4}                         & \textit{II}                           & {2.9}                           & {8.7}                           & {9.1}                          & 7.7                          & {\cellcolor[HTML]{D9D9D9}-85.4} & {\cellcolor[HTML]{D9D9D9}-78.5} & {\cellcolor[HTML]{D9D9D9}-86.8} & \cellcolor[HTML]{D9D9D9}-79.9 \\ \hline
\cellcolor[HTML]{FFFFFF}                    & \cellcolor[HTML]{FFFFFF}\textit{I} & {\cellcolor[HTML]{FFFFFF}0.7}   & {\cellcolor[HTML]{D9D9D9}-1.1}  & {\cellcolor[HTML]{FFFFFF}5.5}  & \cellcolor[HTML]{FFFFFF}12.0    & {\cellcolor[HTML]{D9D9D9}-61.7} & {\cellcolor[HTML]{D9D9D9}-54.6} & {\cellcolor[HTML]{D9D9D9}-67.9} & \cellcolor[HTML]{D9D9D9}-86.9 \\ 
\multirow{-2}{*}{\cellcolor[HTML]{FFFFFF}5} & \cellcolor[HTML]{FFFFFF}\textit{II}   & {\cellcolor[HTML]{D9D9D9}-0.2}  & {\cellcolor[HTML]{FFFFFF}4.2}   & {\cellcolor[HTML]{FFFFFF}9.2}  & \cellcolor[HTML]{FFFFFF}8.1  & {\cellcolor[HTML]{D9D9D9}-56.6} & {\cellcolor[HTML]{D9D9D9}-69.0} & {\cellcolor[HTML]{D9D9D9}-89.0} & \cellcolor[HTML]{D9D9D9}-80.6 \\ \hline
                                            & \cellcolor[HTML]{FFFFFF}\textit{I} & {\cellcolor[HTML]{D9D9D9}-2.3}  & {\cellcolor[HTML]{D9D9D9}-3.1}  & {\cellcolor[HTML]{FFFFFF}3.8}  & \cellcolor[HTML]{D9D9D9}-2.3 & {\cellcolor[HTML]{D9D9D9}-8.4}  & {\cellcolor[HTML]{D9D9D9}-23.5} & {\cellcolor[HTML]{D9D9D9}-44.3} & \cellcolor[HTML]{D9D9D9}-32.9 \\ 
\multirow{-2}{*}{6}                         & \cellcolor[HTML]{FFFFFF}\textit{II}   & {\cellcolor[HTML]{FFFFFF}3.6}   & {\cellcolor[HTML]{FFFFFF}7.2}   & {\cellcolor[HTML]{FFFFFF}7.1}  & \cellcolor[HTML]{FFFFFF}6.8  & {\cellcolor[HTML]{D9D9D9}-72.7} & {\cellcolor[HTML]{D9D9D9}-66.3} & {\cellcolor[HTML]{D9D9D9}-74.7} & \cellcolor[HTML]{D9D9D9}-78.6 \\ \hline
Mean & & {\cellcolor[HTML]{D9D9D9}-0.17} & 1.37 & 3.65 & 2.86 & {\cellcolor[HTML]{D9D9D9}-66.5} & {\cellcolor[HTML]{D9D9D9}-68.8} & {\cellcolor[HTML]{D9D9D9}-76.0} & {\cellcolor[HTML]{D9D9D9}-75.4} \\\hline
\end{tabular}
\end{adjustbox}
\vspace{-0.5cm}
\end{table}

\begin{figure}
\centering
\begin{tabular}{cc}
     \includegraphics[width=0.49\linewidth]{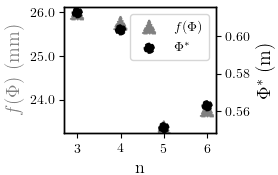} &
     \includegraphics[width=0.49\linewidth]{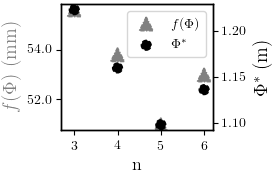} \\
\end{tabular}
\vspace{-0.3cm}
\caption{\small Fitness results for Scenarios (left) \textit{I} and (right) \textit{II}.}
\label{fig:PickPlaceD2}
\vspace{-0.2cm}
\end{figure}

\begin{figure}[!b]
\centering
\includegraphics[width=\linewidth]{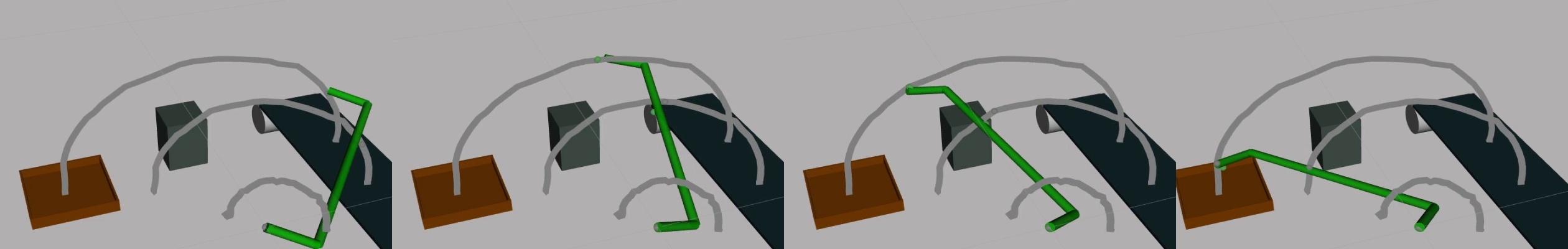}
\vspace{-0.2cm}
\caption{\small A near-optimal 3-DOF robotic arm tracking the human recorded paths and performing the pick-and-place task.}
\label{fig:PickPlaceBest}
\vspace{-0.3cm}
\end{figure}

\begin{figure}[!b]
\centering
\includegraphics[width=\linewidth]{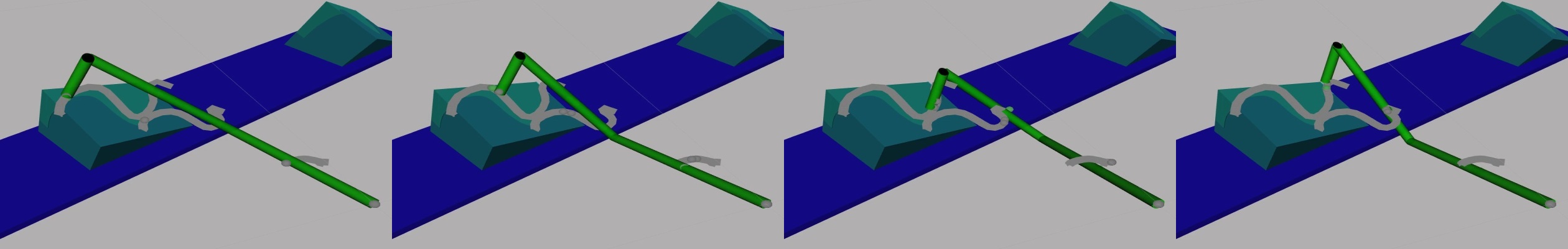}
\vspace{-0.2cm}
\caption{\small A near-optimal 5-DOF robotic arm tracking the human recorded paths and performing a welding task.}
\label{fig:WeldingBest}
\vspace{-0.2cm}
\end{figure}

We seek for $D$ that provides the best performance both in  $\Omega_f$ and CEI. One can choose the value for $D$ individually for a scenario and a specific $n$. However and for generality, we choose the trade-off between CEI and fitness improvements. Hence, we choose to use $D=2$ in which fitness is less compromised while providing significant CEI improvement of 66.5\%. Figure \ref{fig:PickPlaceD2} presents the fitness results for both scenarios with $D=2$ and with regards to $n$. For Scenario \textit{I}, we prefer $n=3$ since the difference in path fitness from larger $n$ is lower than just 4\%. In Scenario \textit{II}, however, the difference is larger as the EE must track the hand orientation. Hence, we choose the minimal fitness design in which $n=5$.  Table \ref{tb:final} presents the parameters of the optimal robots. Figures \ref{fig:PickPlaceBest} and \ref{fig:WeldingBest} show a ROS-Gazebo simulation of the optimal robots tracking recorded paths. Videos of optimal robots can be seen in the supplementary material.


\begin{table}[ht!]
\caption{\small Final parameters of the optimal robotic arms}
\label{tb:final}
\centering
\begin{adjustbox}{width=\linewidth}
\begin{tabular}{c cccc cccc }\toprule
Sc. & $n$ & $\Phi^*$ & $f(\Phi)$ (mm) & $E(\Phi)$ & $i$ & $\alpha_i$ & $a_i$ & $d_{i}$ \\\midrule
\multirow{4}{*}{\textit{I}}  & \multirow{4}{*}{3} & \multirow{4}{*}{0.43} & \multirow{4}{*}{17.59}& \multirow{4}{*}{33.23} & $0$ &0.515 &0 &0  \\   
  &&&& & $1$ &-1.57 &0.11 &0 \\
  &&&& & $2$ &-0.27 &0.5 &0 \\
  &&&& & $EE$ &-1.57 &0 &0.1 \\\hline
\multirow{6}{*}{\textit{II}} & \multirow{6}{*}{5} &
\multirow{6}{*}{0.83} & \multirow{6}{*}{35.83}&\multirow{6}{*}{58.32} & $0$ &$-1.66$ &$0$ &$0$ \\
  &&&& & $1$ &$-0.10$ &$0.09$ &$0.31$ \\
  &&&& & $2$ &$0.86$ &$0$ &$0$ \\
  &&&& & $3$ &$-0.20$ &$0.28$ & $0$ \\
  &&&& & $4$ &$0.66$ &$0$ & $0$ \\
  &&&& & $EE$ &$-0.2$ &$0$ &$0.1$ \\\bottomrule
\end{tabular}
\end{adjustbox}
\vspace{-0.5cm}
\end{table}

\section{Conclusions}

In this work, we have proposed a novel approach for designing the kinematics of a robotic arm to track human demonstrated paths. The design considers the capability of the future robot to accurately track the path of the entire human arm. We have also proposed RA-PSO, a variant of PSO, to efficiently search within the space of design parameters. RA-PSO is shown to be suitable for robot optimization while reducing computational effort. The approach provides a minimal design of a robot arm to do some specific task. The output of the optimization provides also a path in the configuration space to perform the task. The result can save valuable engineering resources in the design phase. In addition, standard and modular hardware can be used to rapidly assemble the outputted optimal kinematics. These may significantly shorten the time-to-operation of the robot. Future work may use the proposed method to optimize both a robotic arm and a computed path to track. 

\bibliographystyle{IEEEtran}
\bibliography{ref}

\begin{thebibliography}{10}
\providecommand{\url}[1]{#1}
\csname url@samestyle\endcsname
\providecommand{\newblock}{\relax}
\providecommand{\bibinfo}[2]{#2}
\providecommand{\BIBentrySTDinterwordspacing}{\spaceskip=0pt\relax}
\providecommand{\BIBentryALTinterwordstretchfactor}{4}
\providecommand{\BIBentryALTinterwordspacing}{\spaceskip=\fontdimen2\font plus
\BIBentryALTinterwordstretchfactor\fontdimen3\font minus
  \fontdimen4\font\relax}
\providecommand{\BIBforeignlanguage}[2]{{%
\expandafter\ifx\csname l@#1\endcsname\relax
\typeout{** WARNING: IEEEtran.bst: No hyphenation pattern has been}%
\typeout{** loaded for the language `#1'. Using the pattern for}%
\typeout{** the default language instead.}%
\else
\language=\csname l@#1\endcsname
\fi
#2}}
\providecommand{\BIBdecl}{\relax}
\BIBdecl

\bibitem{Do2012}
H.~M. {Do}, C.~{Park}, and J.~H. {Kyung}, ``Dual arm robot for packaging and
  assembling of it products,'' in \emph{IEEE International Conference on
  Automation Science and Engineering}, 2012, pp. 1067--1070.

\bibitem{Beschi2019}
M.~Beschi, S.~Mutti, G.~Nicola, M.~Faroni, P.~Magnoni, E.~Villagrossi, and
  N.~Pedrocchi, ``Optimal robot motion planning of redundant robots in
  machining and additive manufacturing applications,'' \emph{Electronics},
  vol.~8, p. 1437, Dec 2019.

\bibitem{Chutima2020}
P.~Chutima, ``A comprehensive review of robotic assembly line balancing
  problem,'' \emph{J. of Intelligent Manuf.}, vol.~8, pp. 1572--8145, 2020.

\bibitem{Cefalo2013}
M.~{Cefalo}, G.~{Oriolo}, and M.~{Vendittelli}, ``Planning safe cyclic motions
  under repetitive task constraints,'' in \emph{IEEE International Conference
  on Robotics and Automation}, 2013, pp. 3807--3812.

\bibitem{Bloch2018}
V.~Bloch, A.~Degani, and A.~Bechar, ``A methodology of orchard architecture
  design for an optimal harvesting robot,'' \emph{Biosystems Engineering}, vol.
  166, pp. 126--137, 2018.

\bibitem{Perrollaz2012}
M.~{Perrollaz}, S.~{Khorbotly}, A.~{Cool}, J.~{Yoder}, and E.~{Baumgartner},
  ``Teachless teach-repeat: Toward vision-based programming of industrial
  robots,'' in \emph{IEEE Inter. Conf. on Rob. \& Auto.}, 2012, pp. 409--414.

\bibitem{Schaal1996}
S.~Schaal, ``Learning from demonstration,'' in \emph{Advances in Neural
  Information Processing Systems}, M.~Mozer, M.~Jordan, and T.~Petsche, Eds.,
  vol.~9.\hskip 1em plus 0.5em minus 0.4em\relax MIT Press, 1996.

\bibitem{Patel2015}
S.~Patel and T.~Sobh, ``Task based synthesis of serial manipulators,''
  \emph{Journal of Advanced Research}, vol.~6, no.~3, pp. 479 -- 492, 2015.

\bibitem{Kennedy1995}
J.~Kennedy and R.~Eberhart, ``Particle swarm optimization,'' in
  \emph{International Conference on Neural Networks}, vol.~4, 1995, pp.
  1942--1948.

\bibitem{Shi2006}
{Shicai Shi}, {Xiaohui Gao}, {Zongwu Xie}, {Fenglei Ni}, {Hong Liu},
  E.~{Kraemer}, G.~{Hirzinger}, and S.~C. {Shi}, ``Development of
  reconfigurable space robot arm,'' in \emph{International Symposium on Systems
  and Control in Aerospace and Astronautics}, 2006, pp. 6 pp.--143.

\bibitem{Vijaykumar1986}
R.~Vijaykumar, K.~Waldron, and M.~Tsai, ``Geometric optimization of serial
  chain manipulator structures for working volume and dexterity,'' \emph{Int.
  Journal of Robotics Research}, vol.~5, no.~2, pp. 91--103, 1986.

\bibitem{Stock2003}
M.~Stock and K.~Miller, ``Optimal kinematic design of spatial parallel
  manipulators: Application to linear delta robot,'' \emph{Journal of
  Mechanical Design}, vol. 125, 06 2003.

\bibitem{Zeiaee2019}
A.~Zeiaee, R.~Soltani-Zarrin, R.~Langari, and R.~Tafreshi, ``Kinematic design
  optimization of an eight degree-of-freedom upper-limb exoskeleton,''
  \emph{Robotica}, vol.~37, no.~12, p. 2073–2086, 2019.

\bibitem{Murray1994}
R.~M. Murray, Z.~Li, and S.~S. Sastry, \emph{A Mathematical Introduction to
  Robotic Manipulation}, 1st~ed.\hskip 1em plus 0.5em minus 0.4em\relax CRC
  Press, Mar. 1994.

\bibitem{You2019}
W.~S. You, Y.~Lee, G.~Kang, H.~Oh, J.~Seo, and H.~Choi, ``Kinematic design
  optimization for anthropomorphic robot hand based on interactivity of
  fingers,'' \emph{Intelligent Service Robotics}, pp. 1--12, 04 2019.

\bibitem{Lum2004}
M.~J.~H. {Lum}, J.~{Rosen}, M.~N. {Sinanan}, and {Hannaford B}, ``Kinematic
  optimization of a spherical mechanism for a minimally invasive surgical
  robot,'' in \emph{IEEE Int. Conf. on Rob. \& Aut.}, 2004, pp. 829--834.

\bibitem{Kuntz2018}
A.~{Kuntz}, C.~{Bowen}, C.~{Baykal}, A.~W. {Mahoney}, P.~L. {Anderson},
  F.~{Maldonado}, R.~J. {Webster}, and R.~{Alterovitz}, ``Kinematic design
  optimization of a parallel surgical robot to maximize anatomical visibility
  via motion planning,'' in \emph{IEEE International Conference on Robotics and
  Automation (ICRA)}, 2018, pp. 926--933.

\bibitem{Rastegar1990}
J.~Rastegar and B.~Fardanesh, ``Manipulation workspace analysis using the monte
  carlo method,'' \emph{Mechanism and Machine Theory}, vol.~25, no.~2, pp. 233
  -- 239, 1990.

\bibitem{Vahrenkamp2012}
N.~{Vahrenkamp}, T.~{Asfour}, G.~{Metta}, G.~{Sandini}, and R.~{Dillmann},
  ``Manipulability analysis,'' in \emph{IEEE-RAS International Conference on
  Humanoid Robots}, 2012, pp. 568--573.

\bibitem{Stocco1998}
L.~Stocco, S.~E. Salcudean, and F.~Sassani, ``Fast constrained global minimax
  optimization of robot parameters,'' \emph{Robotica}, vol.~16, no.~6, p.
  595–605, 1998.

\bibitem{GiladiSintov2020}
C.~Giladi and A.~Sintov, ``Manifold learning for efficient gravitational search
  algorithm,'' \emph{Information Sciences}, vol. 517, pp. 18 -- 36, 2020.

\bibitem{Khatami2002}
S.~{Khatami} and F.~{Sassani}, ``Isotropic design optimization of robotic
  manipulators using a genetic algorithm method,'' in \emph{IEEE Internatinal
  Symposium on Intelligent Control}, 2002, pp. 562--567.

\bibitem{Bryson2016}
J.~Bryson, X.~Jin, and S.~Agrawal, ``Optimal design of cable-driven
  manipulators using particle swarm optimization,'' \emph{ASME. J. Mechanisms
  Robotics}, vol.~8, no.~4, 2016.

\bibitem{Zhu2018}
Y.~Zhu, Z.~Wang, J.~Merel, A.~Rusu, T.~Erez, S.~Cabi, S.~Tunyasuvunakool,
  J.~Kramár, R.~Hadsell, N.~de~Freitas, and N.~Heess, ``Reinforcement and
  imitation learning for diverse visuomotor skills,'' in \emph{Robotics:
  Science and Systems}, 2018.

\bibitem{Ayusawa2017}
K.~Ayusawa and E.~Yoshida, ``Motion retargeting for humanoid robots based on
  simultaneous morphing parameter identification and motion optimization,''
  \emph{IEEE Transactions on Robotics}, vol.~33, no.~6, pp. 1343--1357, 2017.

\bibitem{Laura2020}
L.~Smith, N.~Dhawan, M.~Zhang, P.~Abbeel, and S.~Levine, ``Avid: Learning
  multi-stage tasks via pixel-level translation of human videos,'' in
  \emph{Robotics: Science and Systems (RSS)}, 2020.

\bibitem{Garcia2019}
N.~{García}, J.~{Rosell}, and R.~{Suárez}, ``Motion planning by demonstration
  with human-likeness evaluation for dual-arm robots,'' \emph{IEEE Transactions
  on Systems, Man, and Cybernetics: Systems}, vol.~49, no.~11, pp. 2298--2307,
  2019.

\bibitem{Ceylan2013}
D.~Ceylan, W.~Li, N.~J. Mitra, M.~Agrawala, and M.~Pauly, ``Designing and
  fabricating mechanical automata from mocap sequences,'' \emph{ACM Trans.
  Graph.}, vol.~32, no.~6, 2013.

\bibitem{Coros2013}
S.~Coros, B.~Thomaszewski, G.~Noris, S.~Sueda, M.~Forberg, R.~W. Sumner,
  W.~Matusik, and B.~Bickel, ``Computational design of mechanical characters,''
  \emph{ACM Trans. Graph.}, vol.~32, no.~4, 2013.

\bibitem{Kapusta2016}
A.~Kapusta and C.~C. Kemp, ``Optimization of robot configurations for assistive
  tasks,'' \emph{Georgia Tech Library}, 2016.

\bibitem{Gracia2006}
A.~Perez-Gracia and J.~M. McCarthy, ``Kinematic synthesis of spatial serial
  chains using clifford algebra exponentials,'' \emph{Proceedings of the
  Institution of Mechanical Engineers, Part C: Journal of Mechanical
  Engineering Science}, vol. 220, no.~7, pp. 953--968, 2006.

\bibitem{Shirafuji2019}
S.~{Shirafuji} and J.~{Ota}, ``Kinematic synthesis of a serial robotic
  manipulator by using generalized differential inverse kinematics,''
  \emph{IEEE Transactions on Robotics}, vol.~35, no.~4, pp. 1047--1054, 2019.

\bibitem{rosa2019opytimizer}
D.~R. Gustavo H.~de Rosa and J.~P. Papa, ``Opytimizer: A nature-inspired python
  optimizer,'' 2019.

\bibitem{Dokeroglu2019}
T.~Dokeroglu, E.~Sevinc, T.~Kucukyilmaz, and A.~Cosar, ``A survey on new
  generation metaheuristic algorithms,'' \emph{Computers \& Industrial
  Engineering}, vol. 137, p. 106040, 2019.

\bibitem{Mousakazemi2020}
S.~M.~H. Mousakazemi, ``Computational effort comparison of genetic algorithm
  and particle swarm optimization algorithms for the
  proportional–integral–derivative controller tuning of a pressurized water
  nuclear reactor,'' \emph{Annals of Nuclear Energy}, vol. 136, p. 107019,
  2020.

\end{thebibliography}

\end{document}